\documentclass[letterpaper,10pt,journal,twoside]{IEEEtran}
\usepackage{amsmath,amsfonts}
\usepackage{amssymb}
\usepackage{booktabs}
\usepackage{array}
\usepackage{multirow}
\usepackage{graphicx}
\usepackage{placeins}
\usepackage{cite}
\usepackage{pifont}
\usepackage{xcolor}
\usepackage{fontawesome5}
\usepackage{url}
\usepackage{capt-of}
\usepackage{tikz}
\usetikzlibrary{arrows.meta, positioning}

\newcommand{\cmark}{\textcolor{teal!70!black}{\ding{51}}}
\newcommand{\xmark}{\textcolor{red!70!black}{\ding{55}}}

\definecolor{countgray}{gray}{0.42}
\definecolor{topone}{RGB}{180,35,35}
\definecolor{toptwo}{RGB}{180,125,0}

\begin{document}
\bstctlcite{IEEEauthorControl}

\title{EgoContact: Evaluating Physical Intelligence with Egocentric Video}

\author{Adam Dery\l{}o$^{1,2}$, Alexey Gavryushin$^{3}$, Daniel Cremers$^{2,6}$,
Marc Pollefeys$^{3,4}$, Xi Wang$^{5,3,2,6}$%
\thanks{$^{1}$University of Cambridge, United Kingdom}%
\thanks{$^{2}$Technical University of Munich, Germany}%
\thanks{$^{3}$ETH Zurich, Switzerland}%
\thanks{$^{4}$Microsoft}%
\thanks{$^{5}$Ulm University, Germany}%
\thanks{$^{6}$Munich Center for Machine Learning (MCML), Germany}%
}

\markboth{EgoContact: Evaluating Physical Intelligence with Egocentric Video}{Dery\l{}o \MakeLowercase{\textit{et al.}}: EgoContact: Evaluating Physical Intelligence with Egocentric Video}

\IEEEaftertitletext{%
\vspace{-1em}
\begin{minipage}{\textwidth}
    \centering
    \includegraphics[width=\textwidth]{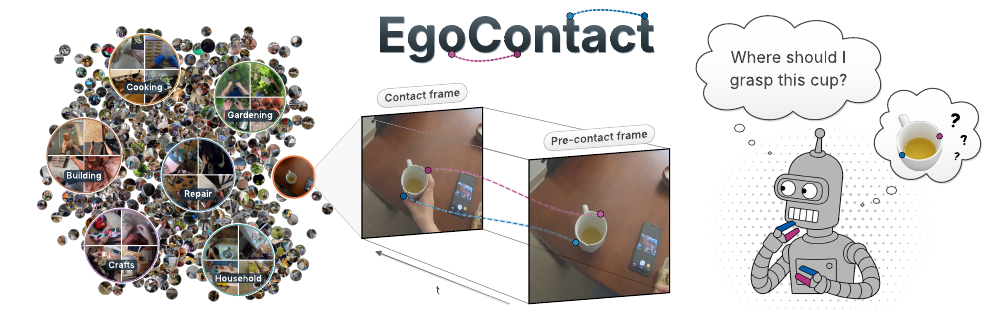}
    \captionof{figure}{\textbf{Overview of EgoContact.} Egocentric video offers a scalable source of interactions with the environment. We use the captured human behavior to probe what the environment affords. Across diverse interactions (\textbf{left}), we annotate objects, actions, and contact points in contact and pre-contact frames (\textbf{middle}), enabling evaluation of where models predict contact for an intended action (\textbf{right}). Project page: \protect\url{https://egocontact.github.io}}
    \label{fig:teaser}
\end{minipage}
\par\bigskip
}

\maketitle

\begin{abstract}
Egocentric video has emerged as one of the most scalable sources of data for learning generalizable manipulation priors. Yet, despite its growing use in pretraining robotic policies, it has been underutilized as a validation tool for physical intelligence, a key component of which is understanding where to make contact with an object to achieve an intended interaction.
We present \textbf{EgoContact}, a large-scale egocentric dataset for evaluating the generalizability and precision of learned manipulation priors through \emph{future contact-point prediction}. EgoContact contains 22.5K manually verified interaction events, each with a pre-contact observation, semantic intent, and pixel-precise contact locations. We find that current methods often struggle to predict precise interaction contacts and sometimes fail to localize the target object in diverse, in-the-wild scenes, even when allowed multiple predictions. Furthermore, we demonstrate the usefulness of EgoContact's annotated interactions for robotic manipulation: simply using EgoContact as the memory bank of an existing affordance-transfer policy improves grasp success from 14.2\% to 31.2\% in simulation and from 16\% to 58\% on a real robot.
\end{abstract}


\section{Introduction}

\IEEEPARstart{R}{obotic} manipulation has advanced rapidly in controlled settings, yet reliable generalization to diverse, in-the-wild scenarios remains elusive. Such generalization calls for more than just passive semantic understanding of objects and available actions, requiring a model capable of translating high-level intent into concrete physical interactions in novel environments. Yet acquiring such physical grounding directly from robot experience is fundamentally difficult to scale, as real-world robot data remains costly and confined to narrow domains.
By contrast, egocentric video offers a compelling source of scalable supervision. Everyday human activity captures a wide range of objects manipulated for diverse purposes in natural environments~\cite{Grauman2021,Damen2018}. Recent work therefore extracts pseudo-labeled interactions from egocentric video to pretrain dexterous and generalizable robotic policies~\cite{Nair2022,Gavryushin2025,Zheng2026}. While pseudo-labeled egocentric data has enabled scaling policy training, it offers no direct way to evaluate the physical knowledge these models actually acquire. %

Evaluating such physical intelligence at scale, however, remains a fundamental challenge. Robot rollouts provide direct evidence, but are expensive, difficult to reproduce, and entangle physical understanding with perception, planning, embodiment, and control. Simulation is repeatable, but restricts the diversity of objects, interactions, and physical conditions and introduces a sim-to-real gap. Existing egocentric benchmarks retain real-world diversity, but primarily model the observed human through actions, narrations, objects, or motion~\cite{Damen2018,Grauman2021,Darkhalil2022,Liu2022}. We instead ask whether human behavior can serve as a reproducible probe of the physical knowledge that makes that behavior possible.

Each observed interaction ties a semantic intent to a concrete realization on an object surface. Knowing that a bottle is graspable, for example, does not specify where the fingers should make contact; recognizing that a drawer will be opened does not identify the handle through which opening can be realized. We therefore focus on a specific component of physical intelligence: the ability to ground an intended action in the object locations that physically support it. Predicting these locations before contact isolates an action-relevant decision without prescribing an embodiment-specific hand pose, trajectory, or controller. This leads to our central question: can current models robustly infer precise, action-conditioned contacts before interaction? Unfortunately, existing benchmarks cannot resolve this question, constrained either by limited spatial precision in their annotations~\cite{Grauman2021} or by insufficient scale and diversity~\cite{Luo2022,Liu2022,Qian2023}.

To close this gap, we introduce \textbf{EgoContact}, a large-scale validation benchmark designed to answer this question. EgoContact contains 22.5K interaction events sampled from unconstrained egocentric video. Each event includes a manually verified contact time, pre-contact observations, semantic intent, object localization, and two pixel-precise contact locations representing the thumb and an opposing finger. These annotations formulate \emph{future contact-point prediction}: given a pre-contact observation and its action context, a model predicts where the subsequent interaction will occur. The task combines the diversity of human video with fixed, manually verified ground truth, enabling repeatable evaluation without requiring robot execution.

Using EgoContact, we evaluate current methods for affordance grounding, affordance transfer, and vision-language reasoning. We find that these methods often struggle to predict precise interaction contacts and sometimes fail to localize the target object, even when allowed multiple predictions. We then examine whether the understanding measured by EgoContact is relevant to robotic manipulation. To investigate this, we replace the memory bank of RAM~\cite{Kuang2024}, an existing affordance-transfer policy, with EgoContact while keeping the rest of the policy fixed. This simple change improves grasp success in both simulation and real-world experiments, demonstrating that contact grounding matters for robotic manipulation.

In summary, we make three contributions:
\begin{itemize}
    \item We introduce \textbf{EgoContact}, a validation dataset that uses human interaction as a scalable probe of precise, action-conditioned physical grounding.
    \item We benchmark a wide range of methods, finding that all struggle to predict precise contact points.
    \item We show that contact grounding matters for manipulation: an EgoContact memory bank improves grasp success in simulation and on a real robot.
\end{itemize}

\section{Related Works}

\paragraph{Egocentric Video Datasets}
Egocentric video datasets such as EPIC-KITCHENS~\cite{Damen2018} and Ego4D~\cite{Grauman2021} offer thousands of hours of unscripted manipulation footage. Benchmarks built on top of these corpora focus primarily on understanding human behavior during hand--object interactions, evaluating tasks such as action recognition~\cite{Damen2018,Perrett2025}, future hand trajectory forecasting~\cite{Liu2022}, and short-term object anticipation~\cite{Grauman2021}. EPIC-Contact~\cite{Bansal2026}, for example, annotates dense 3D meshes and contact surfaces during active manipulation to study human hand--object pose reconstruction. EgoContact departs from this human-centric focus. Rather than modeling human hand kinematics or tracking motion, we use human interactions as physical demonstrations of what the environment affords. By grounding the resulting contact points onto untouched pre-contact frames, we evaluate whether models can infer precise interaction locations from the environment itself.

\paragraph{Robot Learning from Human Behavior}
Egocentric human video has become one of the most scalable supervision sources for robot learning. Earlier work learned reusable visual representations from diverse human video~\cite{Nair2022,Majumdar2023, Gavryushin2025}. More recent efforts train action-conditioned policies at far greater scale~\cite{Zheng2026,Punamiya2026}. Notably, EgoScale~\cite{Zheng2026} reports a log-linear relation between human-data scale and action-prediction loss, which in turn tracks robot performance. Other works~\cite{Punamiya2026,Kareer2024EgoMimic,Yang2025EgoVLA} likewise show that training with egocentric human data improves downstream robot policy performance and generalization. Yet these efforts use egocentric video primarily as a source of training supervision. EgoContact takes this a step further by using observed human interactions to evaluate whether models can ground semantic intent in precise contact locations.

\paragraph{Affordance Grounding}
Given an image and an intended action, affordance grounding methods predict where on an object the interaction should occur. Outputs include heatmaps, part masks, and interaction points~\cite{Luo2022,Qian2023,Qian2024,Ma2025,Yuan2024RoboPoint}. Contacts occupy a useful middle ground: more precise than object- or part-level regions, yet independent of a complete hand pose, trajectory, or controller. Contact-conditioned and hybrid systems similarly factor manipulation into global interaction-point localization followed by embodiment-specific grasp generation or dense control~\cite{Ju2024,Kuang2024,Gavryushin2025,Sundaresan2025HoMeR}. EgoContact isolates the former, testing whether a model can identify the spatially precise contacts needed to guide the latter.
\begin{figure*}[!t]
    \centering
    \begin{minipage}[t]{0.48\textwidth}
        \vspace{0pt}
        \centering
        \includegraphics[width=0.75\linewidth,trim=52bp 52bp 52bp 52bp,clip]{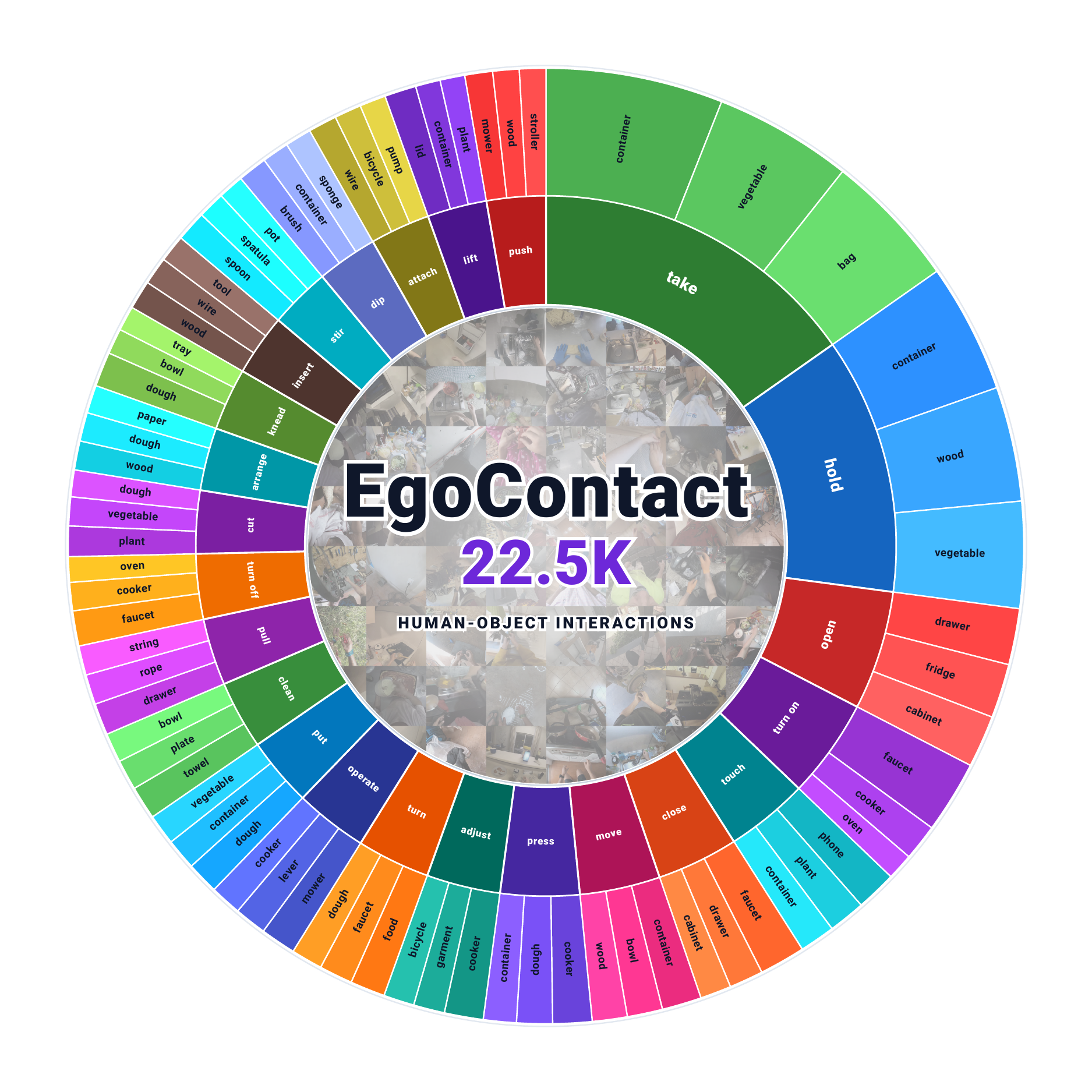}
        \caption{\textbf{Diversity of EgoContact.} The 24 most frequent actions (inner ring) and top three objects per action (outer ring), selected from 22.5K annotated contact events spanning 381 object and 85 action categories.}
        \label{fig:dataset_diversity}
    \end{minipage}\hfill
    \begin{minipage}[t]{0.48\textwidth}
        \vspace{0pt}

\centering
\captionof{table}{\textbf{Comparison of EgoContact and previous benchmarks.}
Existing affordance benchmarks either evaluate on static exocentric images, lack contact-level precision, or provide video heatmaps at very small scale. EgoContact is the first benchmark to combine diverse egocentric video, pre-contact temporal grounding, and manually verified two-point contacts across hundreds of object and action categories at scale. Ego: egocentric view; Temp.: temporally grounded; Obj. and Act.: object and action categories.}
\label{tab:comparison}
\small
\setlength{\tabcolsep}{1pt}
\renewcommand{\arraystretch}{1.65}
\begin{tabular*}{\linewidth}{@{\extracolsep{\fill}}l c c c c l r@{}}
\toprule
Dataset & Ego & Temp. & Obj. & Act. & Annotation & Scale \\
\midrule
AGD20K~\cite{Luo2022}
    & \xmark & \xmark & 50 & 36 & Heatmap & 2,250 \\
3DOI~\cite{Qian2023}
    & \xmark & \xmark & -- & 3 & Contact point & 10,418 \\
Ego4D STA~\cite{Grauman2021}
    & \cmark & \cmark & 110 & 115 & Boxes only & 37,900 \\
\shortstack[l]{Joint Hand\\Motion~\cite{Liu2022}}
    & \cmark & \cmark & 162 & 73 & Heatmap & 401 \\
\midrule
\textbf{EgoContact}
    & \cmark & \cmark & \textbf{381} & \textbf{85} & \textbf{Contact points} & \textbf{22,523} \\
\bottomrule
\end{tabular*}

    \end{minipage}
\end{figure*}

\section{Dataset}

Human behavior reveals what the environment affords. Each successful human--object interaction observed in egocentric video is in essence a display of that object's affordances. While most egocentric datasets are human-centric and seek to model human behavior, we subvert this perspective to focus on what these interactions reveal about the environment being manipulated. With this perspective in mind, we construct EgoContact, a large-scale validation dataset designed to complement existing evaluation frameworks by testing the generalizability of learned manipulation priors across diverse real-world environments.

\subsection{Data Curation}

To evaluate manipulation priors across diverse environments, we source footage %
from Ego4D~\cite{Grauman2021}. While datasets such as EPIC-KITCHENS~\cite{Damen2018} and Egocentric-10K \cite{buildaiegocentric10k2025} provide rich manipulation footage within particular domains, Ego4D captures in-the-wild human-object interactions across a wider variety of settings. We focus our annotation efforts on 37 manipulation-heavy scenarios, including cooking, construction, repair, gardening, crafts, and laboratory work, and exclude scenarios dominated by passive observation, locomotion, or social behavior. This yields 8,626 candidate clips spanning diverse environments, objects, and interaction types (Fig.~\ref{fig:teaser}). %
Across those clips, we use VISOR-HOS~\cite{Darkhalil2022} to automatically detect candidate interaction timestamps. This yields 47,135 candidate events that annotators subsequently verify, adjust, and annotate.

\subsection{Annotations}

We annotate each verified interaction across temporal, spatial, and semantic dimensions.

\textbf{Temporal annotations.} Each event is anchored at the initial contact frame $f_c$, defined as the first frame where contact is established and object manipulation begins. Next, we identify the observation window, an interval of frames $[f_{p_{\mathrm{first}}}, f_{p_{\mathrm{last}}}]$ where the target object is visible and the contact surface is not trivialized by the hand's proximity. We do so by first stepping back at least 0.5\,s from $f_c$ to select $f_{p_{\mathrm{last}}}$ once these criteria are fulfilled, and then stepping back further up to 2.5\,s before $f_c$ to establish $f_{p_{\mathrm{first}}}$. In total, this leaves each event with an uncontacted observation window for anticipation and the contact frame $f_c$ as ground-truth evidence of the object's realized affordance.

\textbf{Spatial annotations.} Having established temporal anchors, we annotate two contact points at $f_c$: the thumb contact and an opposing virtual finger contact. These points are transferred onto $f_{p_{\mathrm{last}}}$, with visibility flags indicating whether a contact lies on the backside of the object. This grounds the realized affordance directly onto the untouched object in $f_{p_{\mathrm{last}}}$. This two-point representation is motivated by the virtual-finger abstraction, in which multiple fingers can act as a functional unit opposing the thumb~\cite{Iberall1987}. MultiGripperGrasp~\cite{Casas2024} further demonstrates successful transfer of parallel-jaw grasp poses to multi-fingered hands in simulation, supporting the use of a compact representation that is not tied to human hand kinematics. Lastly, we also annotate the bounding box of the to-be-interacted object. %

\textbf{Semantic annotations.} Because the same object affords different contacts depending on the action (e.g., slicing with a knife vs.\ handing it over), we describe each interaction in natural language. We normalize these descriptions into canonical verb--noun pairs using Qwen3-VL~\cite{Bai2025} and map them to the Ego4D taxonomy~\cite{Grauman2021} to consolidate paraphrases. This yields action-conditioned queries to test whether models can predict context-dependent contacts on the same object. %

\subsection{Dataset Statistics}

EgoContact comprises 22,523 verified interaction events spanning 381 object categories and 85 canonical action verbs across 37 manipulation scenarios. As shown in Figure~\ref{fig:dataset_diversity}, the action--object distribution is long-tailed: common actions such as taking and holding dominate, while less frequent actions include kneading dough, attaching wires, and operating mowers. The interactions span everyday household tasks, such as opening drawers, turning on faucets, and cleaning plates, as well as outdoor and workshop activities, such as adjusting bicycles and inserting tools. This variety and the in-the-wild nature of the source data allow us to probe the robustness of physical intelligence across diverse methods.

\subsection{Relation to Existing Datasets}

To our knowledge, EgoContact is the first validation dataset for affordance
understanding that combines contact-point precision, large scale, and grounding
in real-world interactions. AGD20K~\cite{Luo2022} and 3DOI~\cite{Qian2023}
provide affordance annotations on static images, without grounding their labels
in observed interactions. Ego4D STA~\cite{Grauman2021} offers temporal grounding
at scale, but its bounding-box annotations support evaluation of target-object
localization rather than precise physical contacts. Joint Hand
Motion~\cite{Liu2022} provides interaction heatmaps grounded in egocentric video,
but is limited to 401 publicly available samples at the time of writing.
Table~\ref{tab:comparison} summarizes these
differences. We omit datasets with automatically derived pseudo-ground-truth
labels, such as HOVA-500K~\cite{Ma2025}, from this comparison. While such labels
can provide scalable training supervision, we have found them to be highly
imprecise at times, limiting their suitability as ground truth for evaluating
physical intelligence. This limitation was one of the motivations behind our
effort.

\section{Experiments}

Evaluating the physical intelligence required for generalizable robotic manipulation presents a trade-off between real-world fidelity, scalability, and reproducibility. On one hand, real-robot rollouts test this capability directly, but are expensive to conduct and difficult to reproduce reliably. On the other hand, evaluation in simulation is repeatable and cost-efficient to scale, but is limited by the realism and diversity of available assets and tasks. We use EgoContact data as the basis for a complementary approach to evaluating knowledge of how to interact with the environment. In doing so, we treat the human behavior depicted in egocentric footage as samples of the affordances of the observed environment. The scale and diversity of such interaction data allow us to probe the physical intelligence of current methods across a broad range of real-world settings. Given such perspective, we aim to answer the following questions:
\begin{enumerate}
    \item Do current models possess physically precise affordance understanding?
    \item And how does it translate to robotic manipulation?
\end{enumerate}

\subsection{RQ1: Do Current Models Possess Physically Precise Affordance Understanding?}

EgoContact enables us to assess the robustness of current methods' affordance understanding at low cost, reproducibly, and at scale. Specifically, we do so through the task of future contact-point estimation.

\subsubsection{Task Formulation}

Given a pre-contact observation $I$ and a semantic or spatial query $q$, a model predicts a ranked list of candidate contact locations $\{\hat{p}_i\}_{i=1}^{K}$ in image space at which contact would realize the specified interaction. Ground truth is obtained by manually annotating the contact locations from the subsequent interaction in the pre-contact observation. In other words, we ask models where they would make contact with an object to carry out the intended action, and compare their predictions with the contact observed in the human interaction. For a common evaluation across methods, we measure prediction accuracy against a single annotated point: the opposing virtual finger contact introduced in the dataset section. The ranked candidates represent alternative predictions of this same contact.

\begin{table}[!t]
\centering
\caption{\textbf{Future contact-point estimation on EgoContact.} Current methods struggle to ground semantic intent into realistic contact points. Notably, no evaluated method beats the naive baseline of predicting the center of the object.}
\label{tab:contact_prediction}
\footnotesize
\setlength{\tabcolsep}{0pt}
\renewcommand{\arraystretch}{1.1}
\begin{tabular*}{\linewidth}{@{\extracolsep{\fill}}l c c c c c c c c@{}}
\toprule
\multirow[c]{2}{*}[-2pt]{Method} & \multicolumn{2}{c}{PiB $\uparrow$} & \multicolumn{2}{c}{mDST $\downarrow$} & \multicolumn{2}{c}{PCK@1\% $\uparrow$} & \multicolumn{2}{c}{PCK@5\% $\uparrow$} \\
\cmidrule(lr){2-3}\cmidrule(lr){4-5}\cmidrule(lr){6-7}\cmidrule(lr){8-9}
 & Top-1 & Top-5 & Top-1 & Top-5 & Top-1 & Top-5 & Top-1 & Top-5 \\
\midrule
\multicolumn{9}{@{}l}{\textit{Naive baseline}} \\
BBox center & 100.0 & 100.0 & 0.04 & 0.04 & 7.36 & 7.36 & 73.78 & 73.78 \\
\midrule
\multicolumn{9}{@{}l}{\textit{Affordance grounding}} \\
GLOVER++ & 21.02 & 28.32 & 0.16 & 0.14 & 0.92 & 2.79 & 14.43 & 20.59 \\
3DOI & 59.96 & 75.33 & 0.10 & 0.06 & 3.49 & 11.95 & 39.30 & 60.62 \\
\midrule
\multicolumn{9}{@{}l}{\textit{Affordance transfer}} \\
RAM & 29.49 & 40.41 & 0.17 & 0.13 & 1.47 & 4.63 & 17.95 & 31.22 \\
Robo-ABC & 33.69 & -- & 0.19 & -- & 1.59 & -- & 18.02 & -- \\
\midrule
\multicolumn{9}{@{}l}{\textit{VLMs}} \\
RoboPoint & 42.18 & 50.25 & 0.13 & 0.11 & 1.19 & 5.04 & 26.47 & 40.78 \\
MolmoPoint & 45.70 & 49.66 & 0.14 & 0.12 & 2.55 & 3.09 & 31.94 & 35.91 \\
Qwen 3.8 27B & 13.74 & 22.56 & 0.25 & 0.21 & 0.34 & 1.73 & 7.78 & 16.30 \\
Gemma 4 31B & 13.68 & 20.71 & 0.22 & 0.19 & 0.45 & 1.25 & 8.27 & 14.54 \\
Muse Spark 1.3 & 57.32 & 79.91 & 0.10 & 0.05 & 3.44 & 11.05 & 42.35 & 71.82 \\
\bottomrule
\end{tabular*}
\end{table}

\begin{figure*}[!t]
\centering
\includegraphics[width=\textwidth]{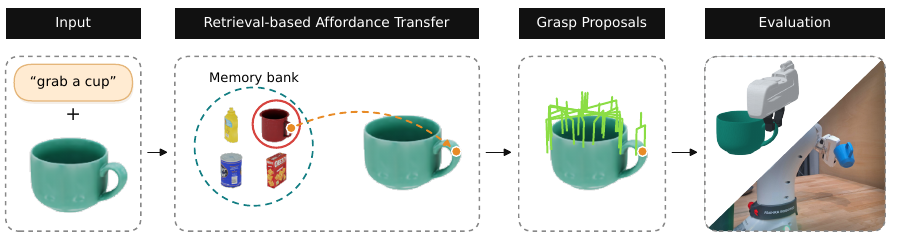}
\caption{\textbf{Grasp evaluation pipeline.} Given an object image and an action query (\textbf{left}), RAM retrieves a relevant interaction from an affordance memory bank of reference images, action descriptions, and contact annotations, then transfers its contact to the target object (\textbf{middle-left}). The predicted contact guides selection among grasp proposals (\textbf{middle-right}), and the selected grasp is evaluated in simulation and on a real robot (\textbf{right}).}
\label{fig:robot_eval_pipeline}
\end{figure*}

\subsubsection{Evaluated Methods}

We evaluate existing methods from three families: affordance grounding,
affordance transfer, and vision-language models. From the affordance grounding
family, we benchmark GLOVER++~\cite{Ma2025} and 3DOI~\cite{Qian2023}; from
affordance transfer, we evaluate RAM~\cite{Kuang2024} and Robo-ABC~\cite{Ju2024}.
Motivated by their use as policy backbones, we evaluate two
point-prediction VLMs, RoboPoint~\cite{Yuan2024RoboPoint} and
MolmoPoint~\cite{AI22026MolmoPoint}, and three general-purpose VLMs,
Qwen 3.8 27B~\cite{Qwen2026Qwen38}, Gemma 4 31B~\cite{Gemma2026Gemma4},
and the state-of-the-art proprietary model Muse Spark 1.3~\cite{Meta2026MuseSpark13}.

All methods receive the observation frame and a semantic instruction, such as
``pick up a cup,'' except 3DOI, which receives the annotated target-object box
center as a spatial prompt. We also use this center as an oracle baseline to
measure contact accuracy from object localization alone.

For heatmap outputs, we extract local peaks with a $3\times3$ maximum filter
and rank them by score, retaining up to ten candidates scoring at least 20\%
of the global maximum. For RAM, we evaluate ranked heatmap peaks, while for Robo-ABC,
we use the contact selected by its retrieval-and-transfer pipeline. A minimal harness collects ranked points via tool calls from general-purpose
VLMs, whereas RoboPoint and MolmoPoint produce native point coordinates.

\subsubsection{Metrics}

We evaluate predictions using Point-in-Box (PiB), mean normalized contact
distance (mDST), and Percentage of Correct Keypoints (PCK) across Top-1
and Top-5 candidates. PiB checks whether any candidate falls inside the
target object bounding box, reflecting coarse localization. Adapting the
distance-to-mask metric from Robo-ABC~\cite{Ju2024} to point contacts, mDST
measures the Euclidean distance from the predicted point to the ground-truth
contact, normalized by the image diagonal, taking the minimum error across
the top-$k$ candidates. PCK measures the fraction of predictions within a
tolerance of 1\% or 5\% of the diagonal. Together, these metrics separate
coarse object localization from fine-grained contact placement.

\subsubsection{Results}

Predicting future contact points from semantic intent remains an open
challenge across all evaluated model families, with no method outperforming
the naive bounding-box center baseline (Table~\ref{tab:contact_prediction}).
Errors stem from two distinct bottlenecks: object localization and fine-grained
contact placement. First, models frequently fail at coarse localization. Low PiB scores indicate
that predictions often miss the target object entirely before any contact
reasoning can take place. Second, even when the object is correctly identified,
precise contact prediction collapses. Among VLMs, Muse Spark 1.3 achieves the
strongest overall performance, reaching 79.91\% Top-5 PiB and 71.82\% PCK@5\%,
yet drops to only 11.05\% at the strict PCK@1\% threshold. Similarly, 3DOI
achieves the highest Top-5 PCK@1\% (11.95\%), but it receives an oracle spatial
prompt that gives away the object location.

Proposing multiple candidates (Top-5) accommodates valid alternative grasp
hypotheses, yet candidate sets still rarely capture the observed interaction
at fine tolerance. Strikingly, the oracle box center achieves 73.78\% PCK@5\%,
surpassing every evaluated model. This demonstrates that performance on loose
distance thresholds is largely driven by coarse object localization rather than
genuine contact understanding. This precision gap directly motivates RQ2:
evaluating whether ground-truth contact precision translates into downstream
robotic manipulation success.

\subsection{RQ2: How does it translate to robotic manipulation?}

EgoContact allows us to evaluate contact-point prediction at scale. But is the
understanding measured by this benchmark relevant to robotic manipulation?
To investigate this, we use RAM~\cite{Kuang2024}, which predicts contacts by
transferring past interactions from a memory bank. We replace its original
memory bank with EgoContact and compare grasp success in simulation and on a
real robot. This setup lets us examine the usefulness of our annotated
interactions while keeping the policy fixed and avoiding changes from fine-tuning.

\subsubsection{Evaluation in Simulation}
We use the MultiGripperGrasp evaluation benchmark in Isaac Sim~\cite{Casas2024}
and its set of 346 GSO and YCB objects. For each object, we render an RGB image
and pass it to RAM with the instruction ``pick up'' and the selected memory
bank (Figure~\ref{fig:robot_eval_pipeline}). We project the predicted contact
into an object point cloud captured from the same viewpoint. GraspGenX~\cite{Han2026GraspGenX}
generates proposals from this cloud, and we select the grasp whose
midpoint is closest to the projected contact. We place a Franka-Panda gripper at the
selected pose and close it. A grasp succeeds if the object remains held under
gravity for at least three seconds.

\begin{figure}[!t]
\centering
\includegraphics[width=\columnwidth]{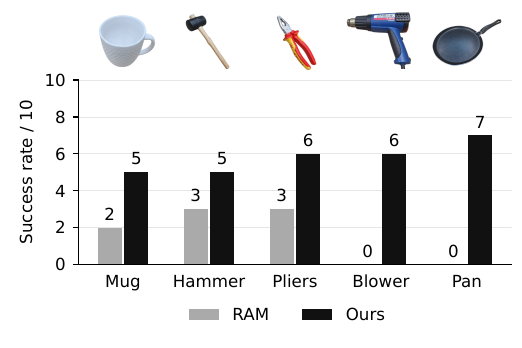}
\caption{Real-robot grasp success on a mug, hammer, pliers, air blower, and pan, with ten trials per object using each memory bank.}
\label{fig:rq2_real_robot}
\par\vspace{4mm}
\includegraphics[width=\columnwidth]{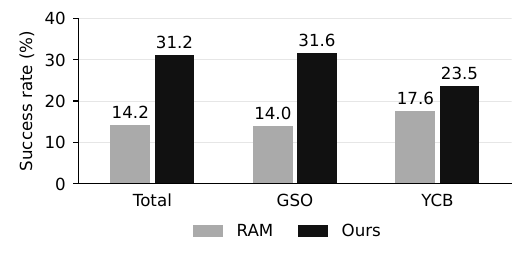}
\caption{Simulation grasp success over 346 objects: 329 from GSO and 17 from YCB. Each memory bank guides one grasp per object.}
\label{fig:rq2_sim_top1}
\end{figure}

\subsubsection{Evaluation on a Real Robot}
We follow RAM's real-robot setup~\cite{Kuang2024}, using a Franka Panda arm,
an RGB-D camera, and AnyGrasp~\cite{Fang2023AnyGrasp} proposals with the same
contact prediction and grasp-selection procedure. We evaluate a mug, hammer,
pliers, air blower, and pan, with ten trials per object and memory bank. After
closing the gripper at the selected grasp, the robot lifts the object. A trial
succeeds if the object remains held after lifting.

\subsubsection{Results}
Using EgoContact improves grasp success in both settings. On the real robot, the improvement
holds across all five objects, including the blower and pan, on which the
original memory bank produces no successful grasps
(Figure~\ref{fig:rq2_real_robot}). Overall, successful trials increase from
8/50 to 29/50. In simulation, success more than doubles relative to RAM's
original memory bank (Figure~\ref{fig:rq2_sim_top1}). These gains are consistent with the intuition that identifying
an actionable region helps select a successful grasp: proposals near a useful
contact are more likely to support the intended interaction. EgoContact's
annotations can thus guide better grasp selection without further training.
The comparison measures the benefit of the memory bank as a whole, including
its interaction coverage and contact annotations.

\section{Conclusion}

We contribute EgoContact, a large-scale egocentric validation dataset for studying
physical intelligence through action-conditioned contact prediction. EgoContact
addresses the need for precise, reliable evaluation across diverse real-world
interactions, which is constrained by the limited scale or coarse annotations of
existing benchmarks and the reliance on automatic pseudo-labels in scalable
egocentric supervision. Using its manually verified contact annotations, we
benchmark a wide range of models and find that all struggle to translate semantic
intent into precise contact predictions that match observed human interactions.
Furthermore, replacing the affordance memory bank of an existing transfer policy
with EgoContact more than doubles grasp success in both simulation and on a real
robot, demonstrating the practical relevance of its annotated interactions. We
hope EgoContact will help the robotics community assess the generalizability of
learned manipulation priors and accelerate progress toward physically grounded
robotic manipulation.

\bibliographystyle{IEEEtran}
\bibliography{main}

\end{document}